# HELFI: a Hebrew-Greek-Finnish Parallel Bible Corpus with Cross-Lingual Morpheme Alignment

Anssi Yli-Jyrä[1,3], Josi Purhonen[1,2], Matti Liljeqvist[3], Arto Antturi[3],
Pekka Nieminen[3], Kari M. Räntilä[3], Valtter Luoto[3]

[1]Department of Digital Humanities, University of Helsinki, 00014, Finland
anssi.yli-jyra@helsinki.fi
[2]Faculty of Theology, University of Helsinki, [3]Raamatun Tietokirja, Aika[media] Oy

**Abstract**
Twenty-five years ago, morphologically aligned Hebrew-Finnish and Greek-Finnish bitexts (texts accompanied by a translation) were constructed manually in order to create an analytical concordance (Luoto et al., 1997) for a Finnish Bible translation. The creators of the bitexts recently secured the publisher's permission to release its fine-grained alignment, but the alignment was still dependent on proprietary, third-party resources such as a copyrighted text edition and proprietary morphological analyses of the source texts. In this paper, we describe a nontrivial editorial process starting from the creation of the original one-purpose database and ending with its reconstruction using only freely available text editions and annotations. This process produced an openly available dataset that contains (i) the source texts and their translations, (ii) the morphological analyses, (iii) the cross-lingual morpheme alignments.

## 1. Introduction

Parallel editions of Bible translations have existed for 1,800 years, but now there is a steadily growing interest to attach a fine-grained alignment to the numerous translations of the Bible and other parallel texts. The interest brings together linguistics, theology, translation studies and language engineering.

Fine-grained manual text alignment produces valuable linked data sets and gold standards for automatic alignment. Tools for producing such manual annotations range from coarse word group aligners (Melamed, 1998b) to ones that align words with confidence levels (Lambert et al., 2005b) and to ones that link various types of subword units (Yli-Jyrä, 1993; Yli-Jyrä, 1995). The obtained gold datasets support the evaluation of alignment algorithms (Mihalcea and Pedersen, 2003; Lambert et al., 2005a; Cysouw et al., 2007) and the development of methods for automated corpus linguistics (Szymanski, 2012).

The aim of this work is to produce an openly shareable, fine-grained alignment for parallel Bibles. Such openly available language resources catalyse research, sharing, linking and tool development.

### 1.1. The Aim

Scientific Bible editions and copyrighted translations are seldom freely shareable. Moreover, manually validated annotations and fine-grained alignments are expensive to produce and thus not typically freely available for research and product development.

The research problem of this paper is to overcome the obstacles to producing an openly shareable, manually aligned fine-grained alignment for a parallel Bible.

### 1.2. The Approach

The paper describes a recently completed process to produce a fine-grained and open morpheme-alignment between the most important 20th century Finnish Bible translation and the most relevant source texts. The whole process covers the years 1991-2020 and consists of two active phases:

**(1) Product Development.** The first phase (1991-1997) added a fine-grained alignment on top of a Bible bitext that contained proprietary components because we had no freely available morphology for the Hebrew source text to rely on. The resulting value-added database was then used to produce, in a few months, a commercial handbook product: a comprehensive analytical concordance of the Finnish Bible translation – *Iso Raamatun Sanahakemisto (IRS)* (Luoto et al., 1997). The concordance was editorially envisioned by Valtter Luoto as an addition to the publisher's long-running series of Bible handbooks (1967–). On the basis of the *work-made-for-hire* rule, the publisher obtained the copyright to the alignment and had the authentic right to decide whether to share the copyrighted alignment with other parties.

**(2) Resource Sharing.** The second phase (2018-2020) began as the publisher made the decision to release the in-house dataset (the alignment) under the CC-BY 4.0 licence. This opening would have been useless without further work on the database. Any remaining proprietary components in the database have now been substituted with open resources by the first two authors. The work resulted in an open resource that contains two bitexts with a fine-grained bilingual alignment. The quality of the whole is comparable with the proprietary database.

### 1.3. The Methodology

In essence, our high-level methodology is a replicable tactic to cope with the delayed availability of open resources. It breaks the production of open annotation resources into the following steps:

1. **Scaffold:** Set up the proprietary text resources as a *scaffold corpus (SC)*.

2. **Addition:** Add valuable annotations to the SC and keep its copyright manageable.
3. **Switchover:** Negotiate an open licence for the copyrighted annotations and replace the proprietary components of SC with open resources.

The **Scaffold** and **Addition** goals were addressed first, during the Product Development phase. Section 3 describes the SC and Section 4 outlines the bilingual annotations. The **Switchover** goal has been addressed recently, during the Resource Sharing phase. In Section 5, we describe how the linked database was detached from proprietary resources and how we implemented its switchover to open resources. Finally, Sections 6 and 7 summarise and evaluate the results, while Section 8 concludes the paper. In the following Section, we describe the background of fine-grained Bible alignment.

## 2. Fine-Grained Bible Alignment

*Fine-grained Bible alignment* is defined as a specialised task where the purpose is to indicate how the shared cross-lingual meaning of the Bible is encoded and aligned between the source and the target texts.

### 2.1. The Bible as a Parallel Text

Parallel texts have existed for a long time, as demonstrated by the Mesha Stele/the Book of Kings (ca 840 BCE), the Behistun Inscription (ca 500 BCE) and the Rosetta Stone (196 BCE).

The *Greek New Testament (GNT)* with more than 2,250 target languages is the most widely translated text in the world. Many of its target languages have several translations. Besides the languages into which the entire GNT has been translated, some parts of it have been translated into an additional 1,800 languages. The *Hebrew Bible* (*HB*, *Tanakh*), also know as the Old Testament, has been translated into more than 700 languages, with the first translation (Septuagint) dating back to 200 BCE.

The *Bible* usually refers to the concatenation of the HB and the GNT. It has influential translations such as the Vulgate (in Latin, CE 405) and the King James Bible (CE 1611). Along with other popular texts like *Le Petit Prince* (360 languages), it is used as a *massively parallel corpus* to research languages (Christodouloupoulos and Steedman, 2015; Jawaid and Zeman, 2010; Tiedemann, 2012; Xia and Yarowsky, 2017; Carlson et al., 2018; Resnik et al., 1998).

### 2.2. Granularity of Alignment

Parallel texts can be aligned at varying levels of granularity (Tiedemann, 2011) and confidence (Lambert et al., 2005b; Holmqvist and Ahrenberg, 2011). The alignment can be carried out either manually by human experts or automatically by algorithms. Manual alignment is time-consuming and expensive but used to produce gold standards for word alignment, while automatic alignment is less accurate but able to handle large amounts of data.

Until recently, the main application of *word alignment* (Brown et al., 1993; Tiedemann, 2011) has been statistical machine translation. Its other applications include lexicon extraction, back-translation, cross-lingual transfer learning, and lexical annotation of the translation. *Linguistically fine-grained alignment* uses lemmas and linguistic tags to align texts in greater detail (Yli-Jyrä, 1993; Yli-Jyrä, 1995; Li et al., 2014). The increased linguistic precision is potentially welcomed in computer-assisted language learning (Nerbonne, 2000), automatic interlinear glossing (Bow et al., 2003; Samardžić et al., 2015), language typology (Cysouw and Wälchli, 2007), and contrastive translation studies (Doval and Nieto, 2019).

### 2.3. Aligned Bibles

The earliest known parallel Bible is that of Origen of Alexandria who compiled the Hexapla (245 CE), a six-column edition of the Hebrew Bible (HB) and its earliest Greek translations. Hundreds of verse-aligned Bible translations are now accessible via Bible.com[1] (1,200 translations), STEP Bible[2] (453 translations), BibleGateway[3] (224 translations), and BibleWebApp[4]. The division into verses was developed for the HB and the GNT by the mid-16th century. The division is relatively stable (±1 verse offset) across translations and it helps to split long sentences (e.g. the sentence in Eph. 1:3-14).

*Strong's Exhaustive Concordance* (Strong, 1890) of an English Bible translation (King James Version) encodes a special kind of manual word alignment. This concordance indicates, for each content word occurrence, the lemma of the corresponding source word with a 4-digit reference number. This number is linked to Strong's Hebrew-Aramaic and Greek lexicons. There are also other concordances with similar source correspondencies (Young, 1879; Åberg, 1982) and several interlinear and reverse interlinear editions of the Bible, all involving an implicit cross-lingual word alignment.

## 3. Scaffold Corpus

We began by preparing the necessary text resources as a SC, without necessarily intending to share it. This contained the Finnish translation, the Hebrew and Greek sources text and the morphological analyses of all texts in the corpus.

### 3.1. Translated Text

In 1992, there were two important, widely used Finnish translations for which there was a commercial interest to develop concordances.

1. the "Church Bible" 1933/38, a relatively literal translation that was already in the public domain, and

---

[1] https://www.bible.com/
[2] https://www.stepbible.org
[3] https://www.biblegateway.com/
[4] http://biblewebapp.com/

2. the 1989-1991 GNT and HB translation proposals by the then-active Bible Translation Committee.

Both translations were linguistically interesting, each in its own way. The proposal followed the principle of dynamic equivalence (Nida and Taber, 2003). However, this principle tends to lead to an inconsistent terminology and less lexical translation correspondencies. As our project's goal was to use lexemes rather than semantic concepts (Louw and Nida, 1989) as the head words for the concordance, we chose to base our research on the older Finnish translation.

In 1991, it was not possible to get hold of an authoritative electronic copy of the the 1933 (HB) / 1938 (GNT) translation. Annoyingly, the circulated digital forms of this translation were inconsistent, and their comparison revealed some 3,000 copying errors and differences in spelling, inflection or punctuation. Some of these seemed to have originated in some printed form of the translation. The first pre-processing task was to compare the variants and to arrive at a reliable consensus between the sources. The official master copy of the translation has now been located in the National Archives of Finland, but not yet consulted.

### 3.2. Source Texts

When a modern publisher translates a book, the author can certify the authenticity and integrity of the original text with a digital signature. This technology has not been available earlier to preserve historical sacred texts or genetic information passed down throughout generations. Biblical manuscripts were not inscribed on a collection of stones, and parchments and papyri were not good for preserving the texts for very long periods of time. Long-term preservation of the texts was dependent on repeated manual copying, correction and verification. Due to the nature of the process, Bible translation is usually based a particular codex or a text-critical edition that aims to indicate the most likely original texts.

The 1933 translation of the HB to Finnish was made using Rudolf Kittel's second edition (1913) of *Biblia Hebraica*. The 1938 translation of the GNT mostly followed a tentative 1913 translation and Eberhard Nestle's latest GNT text edition (Nestle, 1904) but the committee also had access to Erwin Nestle's 13th edition (1927) of *Novum Testamentum Graece*. None of these source text editions was available in digital form when our work started. To set up the source texts in the SC, the project obtained permissions to use the HB text of the *Biblia Hebraica Stuttgartensia (BHS)* (Elliger and Rudolph 1977, 1983) and the GNT text of the 26th edition of *Novum Testamentum Graece (NA26)* (Nestle and Aland, 1979). The minor differences between these text editions and the editions used by the 1933/1938 translators were deemed to have a very limited effect on the source lemmas when we produced the concordance, due to the low number of differences.

### 3.3. Morphological Analyses of Texts

All texts in the SC needed a morphological analysis to support fine-grained alignment.

We obtained the morphological analyses for the BHS edition by licensing the first edition of the *Westminster Hebrew Morphology (WHM) and Lemma Database*. The analysis in this database had been perfected by scholars under the direction of Professor Alan Groves in the Westminster Theological Seminary, based on a draft morphological analysis done by Richard Whitaker (Claremont, Princeton Seminary). Other morphologies for the HB exist today: the Hebrew Bible: Andersen-Forbes[5], Biblia Hebraica Stuttgartensia (Amstelodamensis)[6], and the Open Scriptures Hebrew Bible[7].

The morphological analyses for the NA26 were obtained by licensing Paul Miller's *GNT Database* from the GRAMCORD Institute. Today, many other comparable resources are readily available: the Analytical GNT[8], MorphGNT[9], the Swanson New Testament Greek Morphology[10], various morphologies by Maurice A. Robinson [11], and the Nestle 1904 Morphology[12] etc. For the Finnish translation, we produced a morphological analysis inside the project in collaboration with Professor Kimmo Koskenniemi at the University of Helsinki. His FINTWOL morphological analyser, governed by Lingsoft Oy, was applied in 1992 by Anssi Yli-Jyrä to produce raw morphological analyses for a majority of the word types in the Finnish translation. A few thousand proper names and rarely used words were handled with a post-processing extension to the word-form analyser. The analyser produced, for each token, a set of possible morphological analyses. These analyses were then disambiguated and verified manually by Pekka Nieminen. This produced a golden morphological annotation and lemmatisation for all words in the Finnish translation. Under the collaboration agreement with the University, this annotation was shared automatically with the research community.

## 4. Added Annotation

During the Product Development phase, our priority was to produce the most usable and rich analytical concordance that would nicely complement the already published 10-volume comprehensive encyclopedia of the Bible (Gilbrant et al., 1988). The former 3-volume edition (1967–1972) of the encyclopedia contained a monolingual concordance whose production had involved a lot of manual work. This traditional approach to concordancing was challenged by Arto Antturi and Anssi Yli-Jyrä who innovated in 1991-1992 a new methodology to carry out the editorial work.[13] The new plan was to use computers to produce the (i) keyword-in-context lines and (ii) to extract the source lemmas for each Finnish keyword from the bilingually aligned SC. Human experts were still needed to produce a fine-grained bilingual alignment for the SC.

---

[5] https://www.logos.com/product/25444/
[6] https://github.com/ETCBC/bhsa
[7] https://hb.openscriptures.org/
[8] https://www.agntproject.net/
[9] http://morphgnt.org/
[10] https://www.logos.com/product/179855/
[11] https://github.com/byztxt
[12] http://biblicalhumanities.org/

### 4.1. Alignment Guidelines

In 1992, bitext alignment and statistical machine translation were quite novel tasks in natural language processing. Methods for producing bilingual alignments were evaluated for the first time somewhat later in the ARCADE project (Véronis and Langlais, 2000) but hardly any bitext alignment editors or guidelines were available in the 1990s. In this situation, our team produced a series of internal memos and a graduate school presentation (Yli-Jyrä, 1993) on fine-grained alignment and developed the project's alignment guidelines. These guidelines contained the following key principles:

- alignments are made between groups of tokens that do not need to be continuous
- tokens motivated only by monolingual considerations are aligned to an *epsilon*.
- translations contain the lexical core and the *aux*-linked phrasal periphery (articles and auxiliaries)
- morphological properties of words can be linked separately, with *extractors*, to function words
- the referential content of a pro-form can be linked (via a *pro*-extractor) with a content word without claiming a lexical translation.

All three languages of the SC are synthetic, but the extractors made it possible to link morphological glosses of their words separately from the lexical content. The use of extractors made the alignment more analytical, although this also introduced new kinds of classification errors to the annotations. For Hebrew, the most consistently used extractors included the following:

| %pers | - | person suffix |
| %modus | - | verb modus |
| %tasp | - | tempus-aspect |

Our *aux*-tag and *epsilon* were used very much like the *possible* and *null* confidence levels, developed a decade later by Lambert et al. (2005b).

### 4.2. Alignment Editor

The first editor prototype, *Link*, was bought from IT entrepreneur Sauli Soininen to support Arto Antturi, Kari M. Räntilä and Anssi Yli-Jyrä in their experimentation with the bitext alignment in 1992-1993. Once we had consolidated our alignment guidelines, a more advanced, annotation-aware link editor, *LinkPlus* (Yli-Jyrä, 1995), was developed by Anssi Yli-Jyrä. The editor was implemented in Borland Pascal 7.0 (27th October 1992), and its binaries ran on MS-DOS. Its final 1997 version, constituting 25,000 lines of code, was not just an editor, but an interactive graph database, supporting alternative concordance views to the bitext and provided functions needed to navigate and polish the alignment:

- keyboard-based synchronised bitext navigation
- word alignment across verse/sentence boundaries
- dynamic syntax highlighting for alignment codes
- queries to support validation against guidelines
- concordance view with source language lemmas
- extracting the lines of the printable concordance.

With the *LinkPlus* editor (Yli-Jyrä, 1995), the computer-aided alignment of the whole Bible was carried out in approximately 2.5 man years by theologians Arto Antturi, Kari M. Räntilä, and Matti Liljeqvist. Several review rounds over the alignment were performed.

## 5. Resource Switchover

In 2018, an important step towards the liberation of the aligned Bible was taken by Aikamedia, the publisher and the copyright holder of our concordance: a decision was made to give the in-house alignment data away under the CC-BY 4.0 licence. Right after this decision, the first author started to work on the complete liberation of the aggregated database with his assistant Josi Purhonen. The purpose was to prepare the data for a proper release and to start developing new tools for language learners based on the data.

### 5.1. Locating Open Resources

The base text of the BHS is essentially a copy of *Codex Leningradensis (CL)*, which is in the public domain (they differed only in cantillation and punctuation marks). Their similarity allowed us to replace the proprietary WHM analysis of the BHS with the lemma numbers and morphological analyses provided by *Open Scriptures Hebrew Bible (OSHB)* project under the CC-BY 4.0 license.

As for the GNT, there are currently a few freely available editions: *Nestle 1904*[14], *SBLGNT*[15], *Westcott-Hort*[16], the Byzantine[17] editions, and *Textus Receptus*. *Nestle's 1904* is the closest to the 1938 translation and in the public domain. It has digitised by Diego Renato dos Santos and analysed morphologically by Professor Maurice A. Robinson and Ulrik Sandborg-Petersen.

### 5.2. Synchronising the Text Editions

The SC was built before the UNICODE standard, using *ad hoc* 8-bit transliterations of the non-Roman scripts. The first step in making the alignment compatible with the open resources was to convert the SC resources to the UTF-8 encoding of UNICODE. During the process, the minor punctuation differences between the CL and the BHS involved some extra work.

---

[13] The advice of Krister Lindén and Lauri Carlson is gratefully acknowledged.

[14] https://github.com/biblicalhumanities/Nestle1904
[15] http://sblgnt.com/download/
[16] https://github.com/byztxt/greektext-westcott-hort
[17] https://github.com/byztxt/byzantine-majority-text

Table 1: Tokenisation discrepancies in Hebrew morphologies

| Layer | Example verse | WHM/SC | OSHB | Type of the problem |
|---|---|---|---|---|
| 1 | Mal.3:12.4-5b | כָּל ־ ␣ הַ + גּוֹיִם | כָּל ־ הַ / גּוֹיִם | WHM in SC (our ␣-) versus OSBH (-) |
| 2 | Mal.3:12.1a-1b | וְ + אִשְׁרוּ | וְ / אִשְׁרוּ | Vav-consecutive prefix: WHM (+) versus OSHB (/) |
| 2 | Gen.1:5a-5b | הַ + שָּׁמַיִם | הַ / שָּׁמַיִם | Article: WHM (+) versus OSHB (/) |
| 2 | Ezra.2:61.6a-6b | הַ + קוֹץ | הַקּוֹץ | - OSHB inconsistent |
| 2 | Mal.3:2.11a-11b | כְּ + אֵשׁ | כְּ / אֵשׁ | Prepositions: WHM (+) versus OSHB (/) |
| 2 | 2Kgs.5:8.16 | לָ = מָּה | לָ / מָּה | - WHM inconsistent with preposition ל |
| 2 | Isa.22:18.4a-4b | כַּ + דּוּר | כַּדּוּר | - OSHB inconsistent with preposition כ |
| 2 | 2Chr.26:8.8a-8b | לְ + בוֹא | לָבוֹא | - OSHB inconsistent with ל + the infin. cstr. of בוא |
| 3 | Eccl.4:10.8a-8b | וְ + אִיל = וֹ | וְ / אִי / ל / וֹ | Suffixes: WHM (=, no split) versus OSHB (//) |
| 3 | Mal.3:12.2 | אֶת = כֶם | אֶתְ / כֶם | Suffixes: WHM (=) versus OSBH (/) |

The *Nestle 1904* and *NA26* editions of the GNT contain slightly deviant sets of text-critical readings (there are 700 small differences). Thus some verses or words are found in one GNT edition but are missing from another, but most differences between the text-critical editions were small enough to have no effect on the alignment. Although *Nestle 1904* was closer to what had been translated in 1938, some translated verses were missing from both sources. These were copied from the SC where they had been inserted by the aligners.

### 5.3. Synchronising the Tokenisation

There are significant differences in the way the tokens are segmented in different Hebrew morphologies. Therefore, some manual work was needed to harmonise the tokenisations. The addressed incompatibilities are presented schematically in Table 1.

1. **White-Space Tokenisation.** A common principle in both morphologies is that the main tokenisation boundary corresponds to a white space in the CL. However, contrary to the original WHM analysis, the SC contained a white space (␣) after every *maqqef*-linker. By introducing this white space to the OSHB, a reasonably good overal synchronisation between the SC and the OSHB was achieved.

2. **Hebrew Prefixes.** In Hebrew, prefixes include conjunctions, prepositions and articles. Their boundaries (+) form the second tokenisation layer that assigns alphabetical indices (a,b,c,d) to the subtokens. This layer exists in both morphologies, but some manual work was required to deal with missing or inconsistent prefix boundaries.

3. **Other Affixes.** The OSHB does not make any distinction between prefix or suffix boundaries. In contrast, the WHM does not treat suffixes as subtokens although it marks their boundaries (=). Therefore, some editorial work on the OSHB-based data was also needed to make sure that suffixes did not appear as subtokens. The OSHB also had some analyses that required fixing because the number of lemmas and glosses did not match with the segmentation.

## 6. Results

The research reported in the paper resulted in an open aligned Hebrew-Greek-Finnish Bible and shared knowledge of the methodology applied in preparing it.

### 6.1. Aligned Bible Bitexts

The aligned Bible consists of the HB-Finnish subcorpus (39 books) and the GNT-Finnish subcorpus (27 books). We formatted the released corpora in a new column-oriented format in order to facilitate its use (no conversion to XML format was done at this stage). Tables 2 and 3 show samples from the HB-Finnish and the GNT-Finnish bitexts, respectively:

- Column **token ID** refers to the subtokens of the source language.

- Column **linked IDs** contains links to the source language tokens. A number in parentheses stands for an *aux*-link or translation to an *epsilon*, and a bare dash indicates a lack of source support. The source-side extractors attach to token IDs.

- Column **lemma** indicates the lemma, the corresponding OSHB's enhanced Strong number, and the number of the lexicon entry in the Finnish Analytical Bible Concordance (Luoto et al., eds., 1997–1999, volumes 1-4). The *epsilon*-linked and *aux*-tokens are in parentheses, and a bare dash indicates that there is no Finnish translation. The extractors in the target language side appear in the lemma column.

- Column **morphology** presents the morphological analysis of the token using the *de facto* standard Leipzig glossing rules (Comrie et al., 2008; Lehmann, 1982) but without morpheme boundaries inside subtokens.

- Column **token** shows the tokens as they appear in the text.

- Column **transliteration** shows the source token in a roman-script form according to the academic transliteration standards of the SBL (Collins et al., 2014).

Table 2: The Hebrew-Finnish Alignment of *Psalm 1:1*

| verse | token ID | linked IDs | lemma | morphology | word form | transliteration |
|---|---|---|---|---|---|---|
| ps001:001 | 1 | | -/835/803 | N.MASC.PL.ABS | אַשְׁרֵי־ | ʾašĕrê- |
| ps001:001 | 2a | | -/d/- | ART | הָ | hā |
| ps001:001 | 2b | | -/376/368 | N.MASC.SG.ABS | אִישׁ | ʾîš |
| ps001:001 | 3 | | -/834a/799 | REL | אֲשֶׁר׀ | ʾăšer |
| ps001:001 | 4 | | -/3808/3600 | NEG | לֹא | lōʾ |
| ps001:001 | 5 | | -/1980/1878 | Qal.3MS.PERF | הָלַךְ | hālak |
| ps001:001 | 6a | | -/b/9082 | PREP | בַּ | ba |
| ps001:001 | 6b | | -/6098/5817 | N.FEM.SG.CSTR | עֲצַת | ʿăṣat |
| ps001:001 | 7 | | -/7563/7256 | ADJ.MASC.PL.ABS | רְשָׁעִים | rĕšāʿîm |
| ps001:001 | | 1 | autuas | POS.SG.NOM | Autuas␣ | |
| ps001:001 | | 2a | se | DEM.SG.NOM | se␣ | |
| ps001:001 | | 2b | mies | SG.NOM | mies | |
| ps001:001 | | - | (,) | PUNC | ,␣ | |
| ps001:001 | | 3 | joka | REL.SG.NOM | joka␣ | |
| ps001:001 | | 4 | ei | NEGV.3S | ei␣ | |
| ps001:001 | | 5 | vaeltaa | ACT.PRES.NEG | vaella␣ | |
| ps001:001 | | 7 | jumalaton | PL.GEN | jumalattomain␣ | |
| ps001:001 | | 6b | neuvo | SG.INE | neuvossa␣ | |
| ps001:001 | | 6a | %case | LOC | | |

Table 3: The Greek-Finnish Alignment of *the Epistle to the Hebrews 1:1*

| verse | token ID | linked IDs | lemma | morphology | word form | transliteration |
|---|---|---|---|---|---|---|
| hb001:001 | 1 | | πολυμερῶς/4181/4045 | ADV | Πολυμερῶς | polymerōs |
| hb001:001 | 2 | | καὶ/2532/2515 | CNJ | καί | kai |
| hb001:001 | 3 | | πολυτρόπως/4187/4051 | ADV | πολυτρόπως | polytropōs |
| hb001:001 | 4 | | πάλαι/3819/3685 | ADV | πάλαι | palai |
| hb001:001 | 5 | | ὁ/3588/3455 | ART.NOM.MASC.SG | ὁ | ho |
| hb001:001 | 6 | | θεός/2316/2298 | NOM.MASC.SG | Θεὸς | theos |
| hb001:001 | 7 | | λαλέω/2980&5660/2969 | AOR.ACT.PCP | λαλήσας | lalēsas |
| hb001:001 | | - | (sitten_kuin) | SUB | Sittenkuin␣ | |
| hb001:001 | | (5) 6 | Jumala | SG.NOM | Jumala␣ | |
| hb001:001 | | 4 | muinoin | ADV | muinoin␣ | |
| hb001:001 | | 1 | monesti | ADV | monesti␣ | |
| hb001:001 | | 2 | ja | CNJ | ja␣ | |
| hb001:001 | | 3 | moni | INDEF.SG.ADE | monella␣ | |
| hb001:001 | | 3 | tapa | SG.PTV | tapaa␣ | |
| hb001:001 | | 7 | (olla) | ACT.PAST.3S | oli␣ | |
| hb001:001 | | 7 | puhua | ACT.PCP2.PERF | puhunut␣ | |

### 6.2. Methodological Innovations

Our overall methodology was a replicable template that involved three steps: Scaffold, Addition, and Switchover. This methodology allowed us to cope with the delayed availability of open resources. Each step involved a lot of unique research methodology. This demonstrates the complexity of working with historical documents that have several text editions some of which are under copyright while other editions are in the public domain.

The project developed an alignment editor with unique capacities. If the editor could be updated and ported to modern GUI-based systems, it would still have certain advantages. The alignment guidelines were ground-breaking because they contained ideas that have surfaced only much later: (1) the linguistic distinction between different kinds of links, (2) the extractors, (3) alignment across verse/sentence boundaries, and (4) alignment of discontinuous constructions.

## 7. Evaluation

**Applicability.** Our parallel aligned corpus has several applications. The proprietary version of the dataset was used to produce a concordance handbook. The free version is already being used in order to produce interlinear and reverse interlinear texts to assist in learning of the Bible and biblical languages. The dataset can also be used to evaluate morphological parsers and aligners and to study translation divergencies.

**Market Value.** The alignment-based methodology for producing a bilingual concordance was effective for producing a two-volume handbook in a reasonable amount of time (total 2,000 pages, 300,000 entries). The current work will be a predecessor for further aligned parallel Bibles.

**Alignment Methodology.** The *LinkPlus* editor was one of the pioneering efforts to speed up manal bitext alignment. Since then, several alignment editors have been developed:

- the *PLUG* editor (Merkel, 1999) was one of the earliest alignment editors,
- the *I\*Link* editor (Ahrenberg et al., 2002; Ahrenberg et al., 2003) was interactive but restricted to continuous segments,
- the *Blinker* editor (Melamed, 1998b; Melamed, 1998a) aligned at the level of word groups,
- LinES (Ahrenberg, 2007) was designed for extracting 1-1 correspondencies, and
- the *Alpaco_sp* editor (Lambert et al., 2005b) supported confidence labels.

Although *LinkPlus* was text-based and used file formats that are not used today, it would still be one of the most advanced tools if made available for graphical operating systems.

**Confidence levels.** It is arguable that the *LinkPlus* editor supported three confidence levels (*epsilon*, *aux*, and core links) in links as proposed later (*null, possible, sure*) by Lambert et al. (2005b). Holmqvist and Ahrenberg (2011) argues that such graded links are particularly suited for use in gold standards.

**Granularity.** We are not aware of similar granularity at the level of morphemes for any bitext alignment effort, except in interlinear morphological glossing or in alignment of semantic annotations (Saphra and Lopez, 2015).

**Further research.** In further research, it would be interesting to compare our guidelines with the most recent linguistically designed word alignment guidelines (Li et al., 2008; Li et al., 2014). Another challenge for future research would be to evaluate the inter-annotator agreement under our alignment guidelines. Especially gloss extractors are subject to further validation and correction.

## 8. Conclusion

We have described here an effort to produce an aligned parallel Bible and to convert the produced alignment dataset into a free resource. The free dataset contains two bitexts: the Finnish 1933 translation of the HB according *Codex Leningradensis* and the Finnish 1938 translation based on *Nestle 1904*. The HELFI alignment is released under the CC-BY 4.0 licence on GitHub[18], while other components of the HELFI Corpus[19] are only redistributed under their existing licenses.

## Acknowledgements

The early stages of the project were enabled by collaboration with Lingsoft and the University of Helsinki (UH) and were funded by Aikamedia. The second phase has been enabled by the permission of Aikamedia, and has been supported financially by the UH Faculty of Arts (decision N2/2017 for mobility support and 2018-2019 decisions for research assistance). This report has been written under the Academy of Finland project funding 270354/273457/313478. Its finalisation has also received funding from the European Research Council (ERC) under the European Union's Horizon 2020 research and innovation programme (grant agreement No 771113).

---

[18] https://github.com/amikael/HELFI.
[19] ISLRN 840-665-876-625-0

## Language Resources